\documentclass[10pt,letterpaper]{article}

\usepackage{booktabs} 
\usepackage{subfig}
\usepackage[linesnumbered,algoruled,boxed,lined]{algorithm2e}
\usepackage{graphicx}
\usepackage{amsmath}
\usepackage[a4paper]{geometry}%

\begin{document}

\title{Genie: An Open Box Counterfactual Policy Estimator for Optimizing Sponsored Search Marketplace\footnote{Work is still in Progress}}

\SetKwInOut{Parameter}{Parameters}

\author{
        Murat Ali Bayir\footnote{Corresponding author}\\
        Microsoft Bing Ads\\
				mbayir@microsoft.com
            \and
        Mingsen Xu \\
        Microsoft Bing Ads\\
				mingx@microsoft.com
						\and
				Yaojia Zhu \\
        Microsoft Bing Ads\\
				yaozhu@microsoft.com
					\and
				Yifan Shi \\
        Microsoft Bing Ads\\
				yifanshi@microsoft.com
}

\maketitle

\begin{abstract}
In this paper, we propose an offline counterfactual policy estimation framework called Genie to 
optimize Sponsored Search Marketplace. Genie employs an open box simulation engine with click
calibration model to compute the KPI impact of any modification to the system. From the experimental results on
Bing traffic, we showed that Genie performs better than existing observational approaches that employs
randomized experiments for traffic slices that have frequent policy updates. We also show that Genie can be used to
tune completely new policies efficiently without creating risky randomized experiments due to cold start problem. As time of today,
Genie hosts more than $10000$ optimization jobs yearly which runs more than $30$ Million processing node hours of
big data jobs for Bing Ads. For the last 3 years, Genie has been proven to be the one of the major platforms to
optimize Bing Ads Marketplace due to its reliability under frequent policy changes and its efficiency to minimize
risks in real experiments.
\end{abstract}

\section{Introduction}

Online advertising~\cite{sayedi2014,goldfarb2014} becomes very important as we spend more time on internet based applications like search engines, social network web sites and video streaming services. Sponsored search~\cite{Fain2006,broder2009} represents the largest share of the online advertising~\cite{sayedi2014,zhang2014} where relevant ads (advertisements) are shown to search engine users while they're looking for information. Due to high profitability and user engagement, large companies like Microsoft, Yahoo and Google make significant investments to optimize their sponsored search marketplace.

\begin{figure}[t!]
	\centering
		\includegraphics[width=0.70\textwidth, viewport=40 450 620 700]{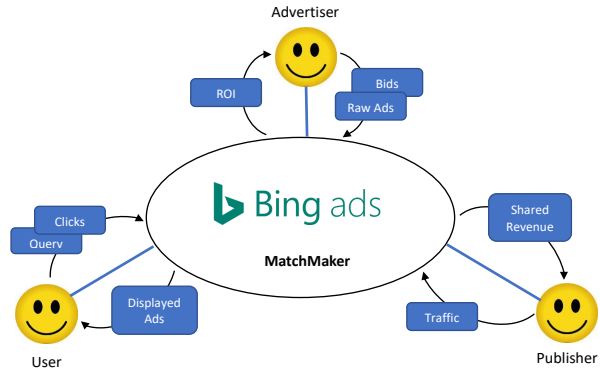}
	\caption{\small Sponsored Search Marketplace \label{fig:marketplace}}
\end{figure}

Sponsored Search Marketplace has four major actors (Figure~\ref{fig:marketplace}) in most general form. Search engines like Bing performs matchmaking role between user, advertiser and publisher. Matchmaker is responsible for delivering relevant ads to the user through the publisher as well as getting bid and raw ads data from advertiser. In some cases, publisher and matchmaker can be same party like Bing. While users interact with the publisher web site, the matchmaker provides ads in back-end and shares the revenue with the publisher. When users are exposed to relevant ads related to their intention, click events lead to return of investment (ROI) for advertisers.

Optimizing Sponsored Search Marketplace is not an easy process since each actor has their own goals and these goals may conflict with each other. The goal of optimization process is to find an equilibrium point which satisfies the constraints coming from all of the parties. The most challenging part of the optimization process is the predicting the Key Performance Indicator (KPI) impact of any new modification to the system which is also knows as counterfactual policy estimation problem~\cite{athey2015,li2015}. 

Counterfactual policy estimation problem attracted many researcher recently within both organic and sponsored search problem context~\cite{bottou2013,li2015,swaminathan15-02}. $A/B$ testing~\cite{kohavi2009,Kohavi13} is known to be most reliable solution to this problem for large scale web applications like sponsored search systems. In $A/B$ testing, users are randomly divided into treatment and control groups and receive ads from two different systems. In this setup, control users receive ads from unmodified application while treatment users receive ads from the application where modification is applied. After that, KPIs from control and treatment traffic slices are compared. While providing very reliable KPI comparison, $A/B$ testing is very difficult to use for many use cases due to end to end deployment requirement and traffic size limitations.

Existing observational approaches like Inverse Propensity Score Weighting~\cite{bottou2013,hirano2001,glynn1989,Rosenfeld17} are another family of counterfactual estimation methods that can be used to perform KPI prediction for policy changes. The idea is to enable subset of real traffic with randomized experiment that represents the modification to the system. Once enough data from randomized traffic is observed with the treatment modification, the offline exploration can be performed on this data to find better operating points. The offline exploration process consumes randomized input to KPI metrics mapping data without running the whole system in open box manner (system is tread as a \textbf{closed box}). While improving traffic limitations compared to $A/B$ testing, existing observational approaches need end to end deployment and exploration in real traffic could be very costly depending on the size of the exploration space. In addition, observational approaches can show large regressions for the data that has bias due to frequent policy updates.

In this paper, we propose a new counterfactual estimation framework called Genie that addresses weak points of previous approaches. Genie utilizes offline simulation engine that performs replaying past event logs in an open box manner and compute KPIs via trained user model from search logs. We claim that Genie has more knowledge on inter dependencies of \textbf{system layer} that impacts the code path from the point where modification applied till to get measurement result. Therefore, replaying the historical data with proper modeling of system inputs (user and/or advertiser signals) becomes very reliable and provide many advantages compared to $A/B$ Testing and existing observational approaches. Apart from KPI prediction reliability, Genie can explore much wider parameter space since it does not require real traffic with modification/exploration cost. Genie can also be leveraged to tune completely new policies where creating initial experiment is very costly due to cold start problem.

Due to its advantages, Genie has been one of the best policy estimation methods for optimizing majority of use cases in Bing Ads that includes adding new policies, large changes in existing policies and any change that is difficult/risky to validate via real traffic. As time of today, Genie hosts more than $10000$ production optimization jobs yearly for Bing Ads Sponsored Search Marketplace. To the best of our knowledge, this paper is the first attempt that propose a log replay based counterfactual estimation system under proper user click behavior modeling. In particular, contributions of this paper are given below:

\begin{itemize}
	\item We modeled Sponsored Search optimization problem as a causal inference problem to predict KPI outcome of any modification to the system. We propose a causal graph with separate input and system layers which could introduce different types of bias in the training data.
	\item We showed that using log replay as an open box simulator for the system subset of causal graph with the proper Machine Learning modeling of input layer yields very reliable KPI predictions.
	
	  \item We also showed that Genie can easily be leveraged to tune completely new policy which is one of most risky use cases for optimizing sponsored search marketplace. We found that KPIs for the best operating points discovered by Genie is statically significant and correlated with KPIs from the real traffic when modification is applied under $A/B$ testing.
\end{itemize}

This paper is organized as follows: Section~\ref{sec:background} discusses background and motivation for this work. The next section summarizes the Related Work. Section~\ref{sec:estimator} introduces the details of Genie counterfactual estimator framework. After that, Section~\ref{sec:experiments} presents our experimental results. Then, we discuss challenges and lessons learned in Section~\ref{sec:challenges}. Finally, we give our conclusions in Section~\ref{sec:conclusions}.

\section{Background and Motivation}
\label{sec:background}

In statistics, causal inference literature defines counterfactual as any change to the system the impact of which is the focus of interest~\cite{athey2015}. In the case of sponsored search, counterfactual could be any policy, parameter or model change in the system that yields a different ad allocation presented to the end user. 

\begin{figure}[t!]
	\centering
		\includegraphics[width=0.8\textwidth, viewport=40 335 620 730]{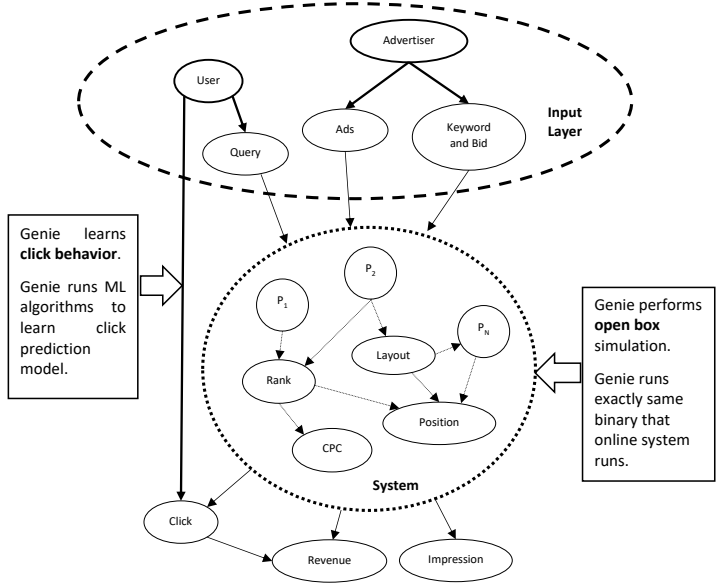}
	\caption{\small Causal Graph for Sponsored Search \label{fig:causal-graph}}
\end{figure}

Similar to the other application contexts, counterfactual estimation for the sponsored search optimization can be modeled as a causal inference problem~\cite{li2015}. For large systems like Bing Ads, the causal graph may contain large system layer with many active policies depending on user and advertiser signals (Figure~\ref{fig:causal-graph}).

\begin{figure}[h!]
	\centering
		\includegraphics[width=0.8\textwidth, viewport=30 535 520 680]{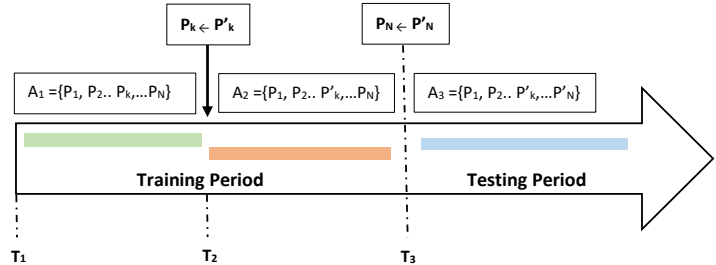}
	\caption{\small Example Tuning with Noise \label{fig:noise}}
\end{figure}

Our main motivation for pushing simulation based counterfactual estimation system comes from the complexity of the causal graph. At a given snapshot, Sponsored Search system may have $N$ active policies $A$ $=$ $\{P_{1}, P_{2}, \ldots, P_{N}\}$ depending on system inputs. For sophisticated systems like Bing Ads, there are frequent updates on the set of active policies (in/out) for each deployment which could potentially create a noise when historical data is used for counterfactual estimation of future KPIs. Consider the following tuning process in Figure~\ref{fig:noise} that optimizes policy $P_{N}$ which is the member of active policies $A_{1}$ in the beginning of the training period. Let say that there is another experiment enabled in all of the traffic slices at time $T_{2}$ which updates the policy $P_{k}$ and changes the distribution of inputs that policy $P_{N}$ consumes. In this particular case, the training time window is needed to be reset from $T_{2}$ and this period should be extended beyond $T_{3}$ to include enough data with the new distribution. However, this could not be practical in many cases due to the frequency of updates and timing constraints.

Unlike existing observational approaches, our simulation based counterfactual estimation can consume all of data for [$T_{1}$, $T_{3}$] time period without resetting tuning process. We propose that KPI impact of any policy can be estimated by replaying training data with the modified policy and using user click behavior model that has tolerable noise. In our policy estimation model, we keep logged advertiser data and user queries fixed while updating click signal based on new page allocation produced by running system layer with new modifications. During the tuning process, the data selection for input layer are randomized by using different seeds to prevent bias due user and advertiser signals.

For the example scenario in Figure~\ref{fig:noise}, if the simulation system has an access to the logged system inputs (user and advertiser signals) and implementation of policy change (usually weeks before creating experiment) in the online binaries for $P_{k'}$, the training data can be simulated with counterfactual policy setting of $A^{'}$ =  $\{P_{1},$ $P_{2},$ $\ldots,$ $P^{'}_{k}$, $\ldots,$ $P^{'}_{N}\}$ for the time period between $T_{1}$ and $T_{3}$.

Apart from supporting tuning scenarios on noisy data given in Figure~\ref{fig:noise}, our simulation platform has other advantages as follows:

\begin{itemize}
	\item Genie could easily used for finding initial settings for randomized experiments to minimize cost of cold start problem.
	\item Applying modification or randomizing system inputs are not possible in certain cases (like bid of advertisers) due to legal issues or traffic capacity. Genie could easily estimate the impact of change in system inputs without real traffic.
	\item Randomized experiments could be very risky in certain cases for new policy changes in terms of user experience. Genie could easily be used for minimizing the risk for these experiments before enabling in real traffic.
\end{itemize}

The reader should also be aware of that some scenarios are difficult to tune both with Genie and existing observational approaches. A typical example is where tuning setup has significant deviations from existing policies/models in real traffic. This type of modifications yield large change in feature distributions of click prediction that Genie is using. For existing observational approaches, tuning requires large exploration space which can significantly regress the user experience. A hypothetical example of this category is that if traffic slice is not tuned for a long time (on the order of several months), there could be large change needed in existing operating points of many policies since marketplace data changed a lot. 

\section{Related Work}
\label{sec:relatedWork}

Counterfactual reasoning and estimation systems~\cite{bottou2013,li2015,swaminathan2015} answer questions like "how would the system performance had been changed if the modification had been applied to the system during log collection?". The idea for these systems is to use online event logs to train models and predict KPIs for new modifications.

$A/B$ testing~\cite{kohavi2007,kohavi2009} is a standard way of evaluating the modifications to the web based systems. In $A/B$ testing setup, users are randomly divided into control and treatment groups. Control group is served by existing binary without any modification whereas treatment group is served by the system where modification is applied. Once enough data is collected from $A/B$ test, control and treatment metrics are compared to test the hypothesis that supports modification to the system.

Observational approaches~\cite{athey2015} like Inverse Propensity Weight Estimation~\cite{bottou2013,hirano2001,glynn1989} were proposed as an alternative to the $A/B$ testing for counterfactual estimation of KPI metrics. Estimation methods like Importance sampling~\cite{bottou2013} can be used in conjunction with online randomization of tunable parameters. Once the logs with randomized parameters are collected, offline exploration can be used to estimate KPI metrics for the proposed parameter settings. As mentioned in previous sections, observational approaches could not be practical in some of tuning scenarios and online explorations should be designed very carefully to minimize negative impact on user experience.~\cite{Schnabel2018}.

Our focus in this paper is to propose efficient counterfactual estimator for practical tuning problems within the context of Sponsored Search System. In particular, we target on tuning scenarios where $A/B$ testing and existing observational approaches can not work efficiently. We split the causal graph for the sponsored search problem into input and system layer and apply different methodologies to predict outcomes. While machine learning methods gives very good results for predicting user click behaviors~\cite{richardson2007,mcmahan2013,Xu10}, we used log replay based \textbf{open box simulator} to estimate the outcome of the sponsored search system. The reader could think of applying similar approaches for policy optimization problem when training data is very noisy due to frequent system changes or when it is not practical to create online experiment with the proposed modification.

\section{Genie Counterfactual Estimator}
\label{sec:estimator}

Genie Counterfactual Estimator corresponds to training job called Auction Simulation that is executed at Microsoft Cosmos Cloud~\cite{chaiken2008}. Auction Simulation job takes many parameters for traffic slice filtration, click model features and user grid that contains counterfactual parameters. Depending on the size of the hyperparameter space, Auction Simulation job can be followed by grid exploration step to improve the efficiency of solution search in hyperparameter space. In the next section, we discuss details of Auction Simulation Job. After that, Grid Exploration step is presented.

\begin{figure}[h!]
	\centering
		\includegraphics[width=1.0\textwidth, viewport=10 325 680 505]{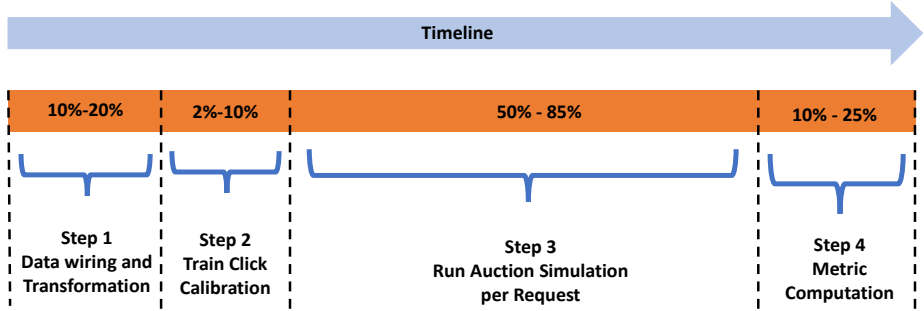}
	\caption{\small Phases of Auction Simulation Job \label{fig:auction-sim}}
\end{figure}

\subsection{Auction Simulation Job}

Auction Simulation Job runs Scope Script on Cosmos~\cite{chaiken2008} to perform the counterfactual estimations. Figure~\ref{fig:auction-sim} presents stages of this step including their relative minimum and maximum computation cost. The details of each phase is discussed in the next subsections.

\subsubsection{Data Wiring and Integration}

The goal of Data Wiring and Integration phase is the reconstructing online library inputs from past event logs. This phase wires logged data from multiple resources that contains ads, auction parameters and page layout data. Then, all of data sources are integrated and transformed into schemas that online library expects. The validation of logged data is performed in this step too.

\subsubsection{Train Click Prediction}
\label{sec:click-predict}

This step is responsible for training model that is used for click prediction. Click prediction is required to re-estimate the click probability of ads in the new page allocation returned from Auction Simulator when counterfactual modification is applied.

Genie click prediction model uses set of $N$ impression feature vectors and click pairs $D$ $=$ $\{$$(x_{1},y_{1})$,...,$(x_{k},y_{k})$, ...,$(x_{N},y_{N})$$\}$ where $\forall$$y_{k}$ $\in$ $\{-1,1\}$ (1: click, -1: no click) as an input. The goal of the click prediction step is to find mapping from each impression vector $x_{k}$ to click probability $\hat{y_{k}}$ where each $\hat{y_{k}}$ $\in$ $[0,1]$.

Due to high runtime performance and high prediction accuracy, majority of Genie estimation jobs are using online version of Bayesian Probit~\cite{graepel2010} for the click prediction problem. 

Bayesian Probit is a generalized linear model with the following posterior distribution of $y$ for given \textbf{$x$} and weight vector of \textbf{$w$} of linear model. 

 \begin{equation}\label{eq1:Posterior}
	p(y|\boldsymbol{x,w}) = {\Phi}{\left(\frac{y.\boldsymbol{w^{T}.x}}{\beta}\right)}
\end{equation}

The probit function $\it{\Phi}(t)$ is the standardized cumulative Gaussian density function that maps the output of the linear model in $[-\infty,\infty]$ to $[0,1]$ with $\beta$ that represent steepness of probit function. 

 \begin{equation}
	{\it{\Phi}}(x) = \int_{-\infty}^{x}N(x,0,1)dx = \int_{-\infty}^{x} \frac{e^{-x^{2}/2}}{\sqrt{2\pi}}dx
\end{equation}

In Bayesian Probit, each impression vector $x_{i}$ contains $L$ features. Each feature $j$ corresponds to $M_{j}$ different discrete bins and represented by a binary vector as below:

 \begin{equation}
	x_{ij} = \left[ {\begin{array}{c} x_{ij(1)} \\ ... \\x_{ij(M_{j})} \end{array}} \right] and \sum_{k=1}^{k=M_{j}}x_{ij(k)} = 1
\end{equation}

Since each feature $j$ of $x_{i}$ corresponds to $M_{j}$ dimensional vector of bins, $x_{i}$ is represented by $\sum_{k=1}^{k=N}M_{k}$ dimensional sparse vector as follows:

 \begin{equation}
	x_{i} = \left[ {\begin{array}{c} x_{i1(1)} \\ ... \\x_{iL(M_{L})} \end{array}} \right] ,{\ }\sum_{j=1}^{j=L}\sum_{k=1}^{k=M_{j}}x_{ij(k)} = L
\end{equation}

The probit model stores as a vector of gaussian distributions for each bin that represents the weights of generalized linear model as follows:

 \begin{equation}
	\textbf{}{w} = (\boldsymbol{\mu}, \boldsymbol{\sigma}),  \boldsymbol{\mu} = \left[ {\begin{array}{c} \mu_{11} \\ ... \\ \mu_{L(M_{L})} \end{array}} \right] ,{\ }\boldsymbol{\sigma}^{2} = \left[ {\begin{array}{c} \sigma^{2}_{11} \\ ... \\ \sigma^{2}_{L(M_{L})} \end{array}} \right]
\end{equation}

During the model training, each request data is converted to a set of impression and click signal \{-1,1\} features on Cosmos. Each impression vector and click signal pair is only scanned one time in an online manner. During training data scan, the mean and variance of Gaussian distributions for matched weight bins are updated via backward message passing~\cite{graepel2010} to maximize the posterior probability in Equation~\ref{eq1:Posterior}. Once the model is trained, the inferred model is sent to downstream components of the scope script. For an evaluation of given feature vector $\boldsymbol{x}$, the Probit model accumulates variance and mean of those matched feature bins as follow:

 \begin{equation}
	\sigma^2={\boldsymbol{x}^T}{\boldsymbol{\sigma}^2} ,{\ } {\mu}={\boldsymbol{x}^T}{\boldsymbol{\mu}}
\end{equation}

Then, predicted click probability is calculated as $\it{\Phi}\,(\mu/\sqrt{\sigma^2 + \beta^2})$ where $\it{\Phi}$ is the standardized cumulative Gaussian density function.

Apart from online Bayesian Probit, Genie also supports Click Prediction model based on Gradient Tree Boosting~\cite{chen2016}. Tree Boosting is a state of art method that is widely used in many real world applications including search result ranking and click through rate prediction.

Tree Boosting model implemented in Genie works with both continuous and categorical features. Therefore, it does not require costly feature engineering process to create binned feature space like in Bayesian Probit. Tree boosting model has $L$ regression trees and each tree $T_{k}$ ($k$ $\leq$ $L$) is traversed from top to down to find a particular leaf node for a feature vector $\textbf{x}$ via conditions on intermediate nodes. The weight of discovered leaf $w_{k}$ is used as a weight for current tree $T_{k}$. This process is repeated for all of the trees. After that, a sigmoid function is applied to the sum of weights to create final click prediction value between $[0,1]$ as below:

\begin{equation}
	y_{i} = \sigma(\sum_{k=1}^{k=N}w_{k}),{\ } \sigma(x) = \frac{1}{1 + e^{-x}}
\end{equation}

During the training phase, each request data is converted to a set of features that is very similar to the one for Bayesian Probit. After that, the training data is scanned $L$ times in batch manner to train tree ensemble. Inspired from Friedman et al.~\cite{friedman2001}, the following log loss objective function on prediction error is propagated in each iteration $t$ to create new trees:

 \begin{equation}\label{eq:loss-boosting}
	Loss^{t}=\sum_{i=1}^{i=N}\psi(y_{i},\hat{y_{i}}^{t})=\sum_{i=1}^{i=N}\left[-y_{i}\log{\hat{y_{i}}^{t}} - (1-y_{i})\log({1-\hat{y_{i}}^{t}})\right]
\end{equation}

 \begin{equation}\label{eq:estimation-boosting}
	\hat{y_{i}}^{t}=\sigma{\left(\sum_{k=1}^{k=t}w^{k}\right)}
\end{equation}

While constructing the single tree in iteration $t$, the Equation~\ref{eq:loss-boosting}-\ref{eq:estimation-boosting} are used to produce target value for each data sample that corresponds to the gradient of the error function given in Equation~\ref{eq:gradient}.

 \begin{equation}\label{eq:gradient}
	\bar{y_{i}}^{t} = \left[- \frac{\partial{\psi(y_{i},\hat{y_{i}}^{t-1})}} {\partial{\hat{y_{i}}^{t-1}}} \right]
\end{equation}

After that, the split gain function is derived from least squares on target values to select feature with highest gain to create new branches in current iteration. The split process is done iteratively in greedy manner starting from feature that has highest split gain. 

From the counterfactual policy estimation perspective, both methods has different trade offs. While tree boosting eliminates the process for costly feature engineering, it does not fit into traditional Map Reduce paradigm well compared to Bayesian Probit since it requires multiple disk read and write for each iteration. On the other hand, we found that Tree Boosting yields models with higher accuracy in most common tuning scenarios as mentioned in Experimental Results Section. As time of today, Genie counterfactual estimator keep both methods and user config defines the one that is going to be applied based on efficiency and quality trade offs.

\subsubsection{Auction Simulation}

This phase executes Bing Ads online libraries with the modified version of the reconstructed inputs per request. The offline replay process of the online binary with same or modified inputs is called \textbf{open box auction simulation}.

The Auction Simulaton contains a wrapper layer which is responsible for reading all of the inputs and invoking the online library. Simulation wrapper reads the counterfactual grid points, click prediction model and auction data. Then, each counterfactual grid point is converted into modifier method. The Modifier method mutates the auction data in place and returns the restorer function for auction data and object representation of the current grid point. Once the online library is called with the mutated input, the page allocation for the mutated input is generated. After that, click predict model is applied to re-evaluate click probability of ads in the page allocation. Finally, KPI metrics for current settings are computed from updated page allocation. The pseudo code for the simulation logic is given in Algorithm 1.

\begin{algorithm}
  \KwIn{Auction Data: $A$, Grid: $G$, Click Model: $C$}
  \KwOut{List of (Setting, KPI) pairs as $KPIs$}
	$M$ $\leftarrow$ $G$.GenerateModifiers($A$)\\
	$KPIs$ $\leftarrow$ \{ \} // Initialize the output.\\
	\ForEach{$M_{i}$ $\in$ $M$}{
	    // Modify the input and get (restorer, setting)\\
			($R_{i}$, $S_{i}$) $\leftarrow$ $M_{i}$.Modify($A$)\\
			$P_{i}$ $\leftarrow$ OnlineLibrary($A$) // Create $i^{th}$ page allocation\\
			C.Predict($P_{i}$) // Adjust click probabilities\\
			$KPI_{i}$ $\leftarrow$ GetKPI($P_{i}$)\\
			$KPIs$ $\leftarrow$ $KPIs$ $\cup$ $(S_{i}, KPI_{i})$\\
			$R_{i}$.Restore($A$) // Restore the input to original value\\
	 }
\caption{Simulation Algorithm}\label{algo:simulation}
\end{algorithm}

The key part of the Algorithm 1 is the simulation of the online library (line 6) for running an auction. There are many ways to design an online library that runs an auction~\cite{Goel2015,Varian2007,varian2014} in real time. Here we only discuss the general form described in Varian et al~\cite{Varian2007} based on Generalized Second Price Auction. The subsequent auction logic described in this paper is consistent with definitions in Variant et al~\cite{Varian2007}. However, our framework can run with any type of auction mechanism as long as the interface remains same.

\begin{algorithm}[h!]
  \KwIn{Available page templates $P$, ads $A$}
  \KwOut{Best page allocation $P_{max}$}
	Sort($A$) // Sort ads based on rank score\\
	$P_{max}$ $\leftarrow$ null // initialize best allocation\\
	\ForEach{$P_{i}$ $\in$ $P$}{
		\ForEach{$block_{j}$ $\in$ $P_{i}$}{
			\For{$k \leftarrow 0$ \KwTo $block_{j}.Size()$}{
				$winner$ $\leftarrow$ FirstEligible($A$, $block_{j}$)\\
				$runnerup$ $\leftarrow$ NextEligible($A$, $winner$, $block_{j}$)\\
			  $block_{j}[k]$.$Ad$ $\leftarrow$ $winner$ \\
				$block_{j}[k]$.$PricingScore$ $\leftarrow$ $runnerup$.$RankScore$
			}
		}
		\uIf{$P_{max}$ $=$ null \textbf{or} $P_{max}$.Utility $<$ $P_{i}$.Utility}
			{$P_{max}$ $\leftarrow$ $P_{i}$}
	 }
\caption{Allocation Algorithm for GSP}\label{algo:allocation}
\end{algorithm}

In general form, the auction system can get 3 sets of input data from simulation wrapper. These inputs can contain basic ad data, PClick scores and other ad specific metadata. In addition to data inputs, auction layer can also get settings for different policies and candidate page layouts to select. Following ~\cite{bottou2013,Varian2007}, the rank score or the utility of $ad_{i}$ can depend on bid $b_{i}$ of current advertiser and ad dependent click quality score $q_{i}$. The utility $u_{i}$ of $ad_{i}$ can be defined as $u_{i}$ $=$ $b_{i}$$q_{i}$ where $q_{i}$ contains click probability $p_{i}$ of current ad. Once the utility computation for each ad is done, ads can be placed on page layouts starting from most significant position to the least significant in a greedy manner. For each available position in the page layout, the winner ad and runnerup can be determined based on their rank scores. The runnerup can be used for determining pricing score for the current position which is required for calculating the cost per click (CPC) based on Generalized Second Price Auction~\cite{Varian2007}. Finally, the page allocation with the maximum utility can be selected as a best allocation. he high level pseudo code for this simple allocation logic is given in Algorithm 2.

\subsubsection{KPI Computation}

KPI computation phase computes metrics of interest for each counterfactual grid setting. Aggregated metrics are used in decision process to update the operating points of several policies in the real serving environment. In this phase, metric data for each request and grid point pair are converted into data cubes. Each data cube contains $n$ dimensions with many datacube cells. A data cube cell contains KPI metrics that is pivoted by $n$ dimensional key. Once the data cube for each request and grid point is created, they are aggregated recursively with sum operation over all requests to produce final metrics.

\subsection{Grid Exploration}
\label{sec:grid}

Some of the tuning scenarios need to explore very large hyperparemeter space that contains more than $15$-$20$ policy parameters. As a result, these tuning jobs requires significant amount of computing resources to run simulation based approach. For an efficient hyperparameter space search, the Grid Exploration step can be executed between sequential tuning jobs. In this setup, grid exploration step reads simulated data points from former job and recommend better operating points for latter jobs based on the model trained from simulated points of former job.

In an nutshell, Grid Exploration step learns a regression model that maps set of hyperparameters $X$ to set of real valued KPI deltas ${\Delta}Y$. KPI delta refers to normalized difference between KPI that is estimated by applying grid point (treatment) and KPI that is estimated from logs without any modification (baseline). This step learns a separate instance of regression model for each metric that tuning user is interested. For simplicity we assume that each ${\Delta}y_{i}$ $\in$ ${\Delta}Y$ is one dimensional real value and each $x_{i}$ $\in$ $X$ is a $k$ dimensional vector. For a given data set $D$ $=$ $\{$$(x_{1},{\Delta}y_{1})$,...,$(x_{k},{\Delta}y_{k})$,...,$(x_{N},{\Delta}y_{N})$$\}$ the Grid Exploration step can learn two types of regression model.

In linear regression case, the least squares approach is used to minimize the following sum which yields solution for coefficients $\beta$ $=$ $($$\beta_{0}$, $\beta_{1}$,...,$\beta_{k}$$)$:

\begin{equation}
  Min_{\beta} \left\{ \sum_{i=0}^{N}\left({\Delta}y_{i} - \sum_{j=1}^{k}{\beta_{j}}x_{ij} - \beta_{0}\right)^{2} \right\}
\end{equation}

For the ridge regression, the extra term is added to the error function to shrink the linear coeffients $\beta$ with shrinkage factor $\lambda$ to avoid potential overfitting:

\begin{equation}
  Min_{\beta} \left\{ \sum_{i=0}^{N}\left({\Delta}y_{i} - \sum_{j=1}^{k}{\beta_{j}}x_{ij} - \beta_{0}\right)^{2} + \lambda{\sum_{j=0}^{k}{\beta_{j}^{2}}} \right\}
\end{equation}

\begin{algorithm}
  \KwIn{Hyperparameters: $X$, KPI deltas: ${\Delta}Y$, Batches: $B$}
	\KwIn{Population size $P$, Solution size: $k$, Objective: $F_{obj}$}
  \KwOut{Best solution set: $X'$ and ${\Delta}Y'$}
	Model = Train($X$, $Y$)\\
	$X'$ $\leftarrow$ $X$, ${\Delta}Y'$ $\leftarrow$ ${\Delta}Y$\\
	\For{$i \leftarrow 1$ \KwTo $B$}{
		$X_{i}$ $\leftarrow$ Explore($X'$, $P$) // Create new solution set\\
		${\Delta}Y_{i}$ $\leftarrow$ Model.Predict($X_{i}$)\\
		// Select best $P$ solution from current union previous.
		${\Delta}Y'$, $X'$  $\leftarrow$ Select($X'$ $\cup$ $X_{i}$, ${\Delta}Y_{i}$ $\cup$ ${\Delta}Y'$, $P$)\\
	 }
	// Select the top $k$ solution among ${\Delta}Y'$\\
	($X'$, ${\Delta}Y'$) $\leftarrow$ Top($X'$, ${\Delta}Y'$, $k$, $F_{obj}$)
\caption{Optimization Algorithm}\label{algo:optimization}
\end{algorithm}

Once the model is trained, the iterative optimization algorithm similar to hill climbing is executed with the user supplied parameters. The details of the optimization process is given in Algorithm 2. In each iteration, a temporary solution set is created by exploring neighbours of the current solution set. The most common exploration logic is selecting random values from $[min, max]$ range of random subset of each hyperparameter dimensions based on uniform distribution. Then, the best $P$ (population size) solutions from the union of current and previous step are chosen to finalize the solution set in the current step. The iteration stops after certain number of executions and top $k$ solutions from the solution set are selected as final result.

\section{Experimental Results}
\label{sec:experiments}

In this section, we provide our experimental results on Bing serving traffic. We first present experimental results on the the quality and efficiency of click prediction, auction simulation and grid exploration steps. After that, we compare the counterfactual estimation accuracy of Genie with an Importance Sampling based estimator on one of most noisy and largest traffic group in Bing. Finally, we present counterfactaul estimation accuracy results of Genie on one of the most difficult use cases that requires tuning new policy.

\subsection{Click Prediction Performance}


In the click prediction experiments, we used $10$ million sample of data that spans a week time period from Bing PC traffic of January 2018. For the experimental setup, the size of training data is gradually increased from $100$K to $3.2$M. In each data point, the subset of data excluding the training set is used for testing.

\begin{figure*}[ht!]
  \begin{center}
  \begin{minipage}{0.5\linewidth}%
    \centering
		\subfloat{\includegraphics[width=1.0\textwidth, viewport=130 530 520 705]{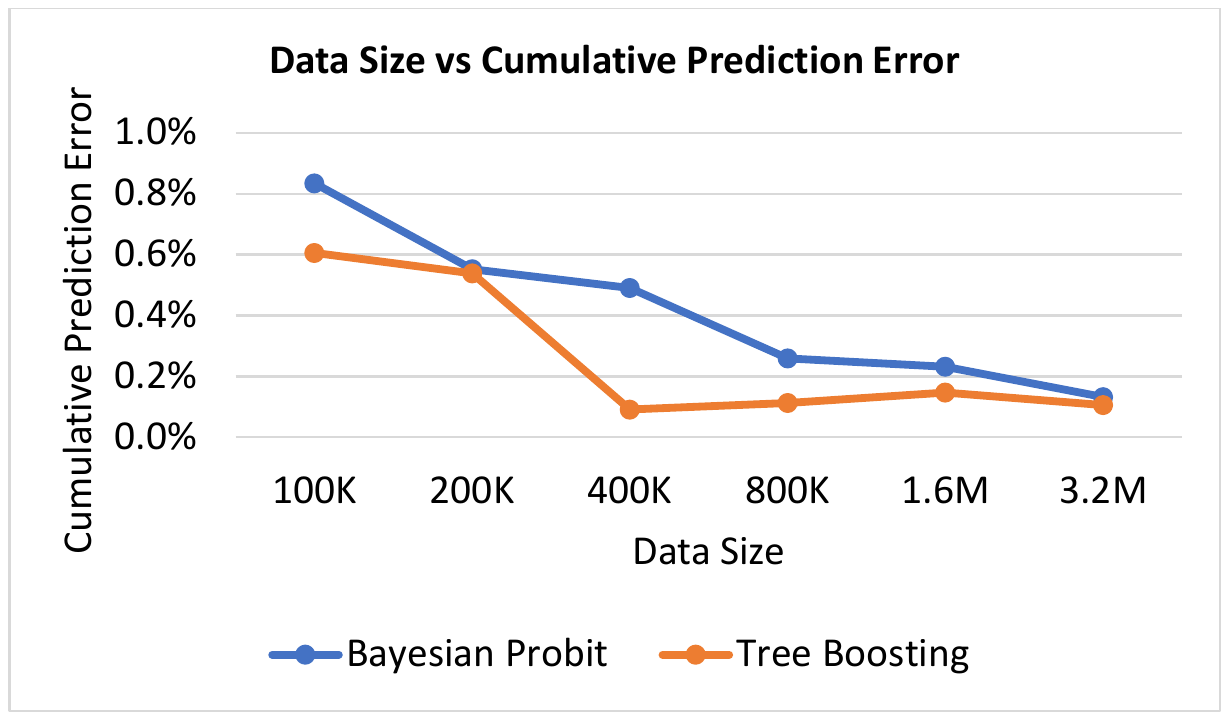}}%
	  \caption{\small Cumulative Prediction Error \label{fig:cc-cumulative}}%
  \end{minipage}%
	\begin{minipage}{0.5\linewidth}%
    \centering
		\subfloat{\includegraphics[width=1.0\textwidth, viewport=130 530 520 705]{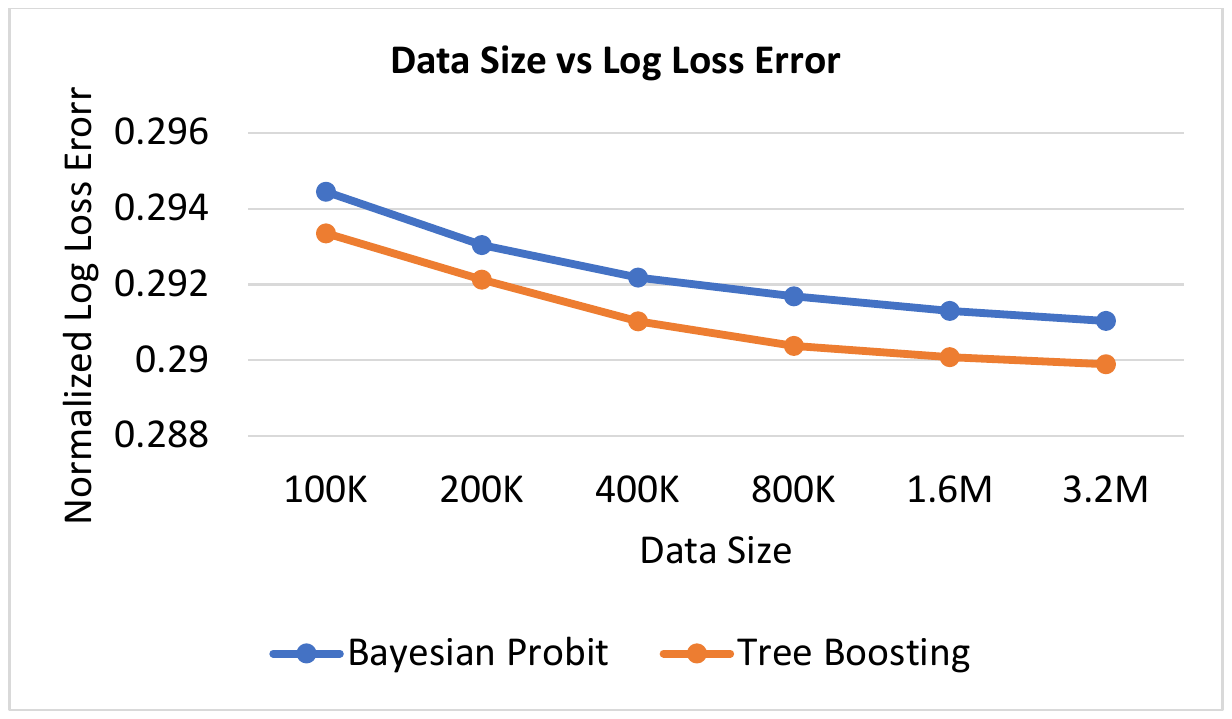}}%
	  \caption{\small Log Loss Error \label{fig:cc-log-loss}}%
  \end{minipage}%
  \end{center}
\end{figure*}

Figure~\ref{fig:cc-cumulative} presents the cumulative error for click prediction models with respect to Equation~\ref{eq:cumulative} where $N$ is the number of test impressions.

\begin{equation}
\label{eq:cumulative}
  E_{cumulative} =  \frac{\left|{\sum}_{i=1}^{N}y_{i} - {\sum}_{i=1}^{N}\hat{y_{i}}\right|}{{\sum}_{i=1}^{N}y_{i}}
\end{equation}

For the cumulative error, we observed that Tree Boosting is slightly better than Bayesian Probit when the data size is smaller than $3M$ samples. Another observation is that when data size becomes larger than $3$M samples, the error rate is getting stable and converges for homogenous traffic like Bing PC. This behavior is very similar to other homogenous traffic slices that are tuned by Genie.

\begin{figure}[h!]
	\centering
		\includegraphics[width=0.5\textwidth, viewport=130 530 520 705]{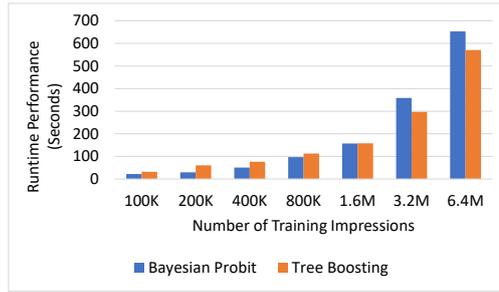}%
	\caption{\small Runtime Comparision \label{fig:cc-perf}}%
\end{figure}

While the cumulative error gives general idea for click prediction quality on the aggregated level, it does not reflect the success of individual impression predictions. The normalized Log Loss error in Equation~\ref{eq:log-loss} gives better estimation of impression level prediction accuracy. In the second experiment, we compared the Log Loss error of Bayesian Probit and Tree Boosting (Figure~\ref{fig:cc-log-loss}). We observe that Tree Boosting is slightly better than Bayesian Probit in terms of Log Loss Error. The convergence behavior is very similar to cumulative prediction error after $3$M samples. 

\begin{equation}
\label{eq:log-loss}
  E_{logloss} =  \frac{1}{N}\sum_{i=1}^{i=N}\psi(y_{i},\hat{y_{i}})
\end{equation}

In the last experiment, we compare the runtime performance of the model training part of Bayesian Probit and Tree Boosting. The training part for the Bayesian Probit runs on cosmos and integrated into Auction Simulation Job. The training job for Tree Boosting is running in a single box on Microsoft Internal Workflow Platform outside the Auction Simulation Job. Figure~\ref{fig:cc-perf} presents the runtime comparison of training jobs for these two models. We observe that the training part of both model is very fast and have similar performance. However, the current Tree Boosting implementation is not compatible with Cosmos and running outside the cosmos data cluster. Therefore, the significant amount of time (10 min - 2 hours) is spent on transferring training data between data centers (Figure~\ref{fig:cc-perf} excludes this data). Because of this, Tree Boosting is only preferred for cases when significant feature engineering is involved (non continuous features) or noisy traffic slices in terms of user behavior and feature space (traffic types where Bing is not publisher).

\begin{figure*}[ht!]
  \begin{center}
  \begin{minipage}{0.5\linewidth}%
    \centering
		\subfloat{\includegraphics[width=1.0\textwidth, viewport=130 535 510 710]{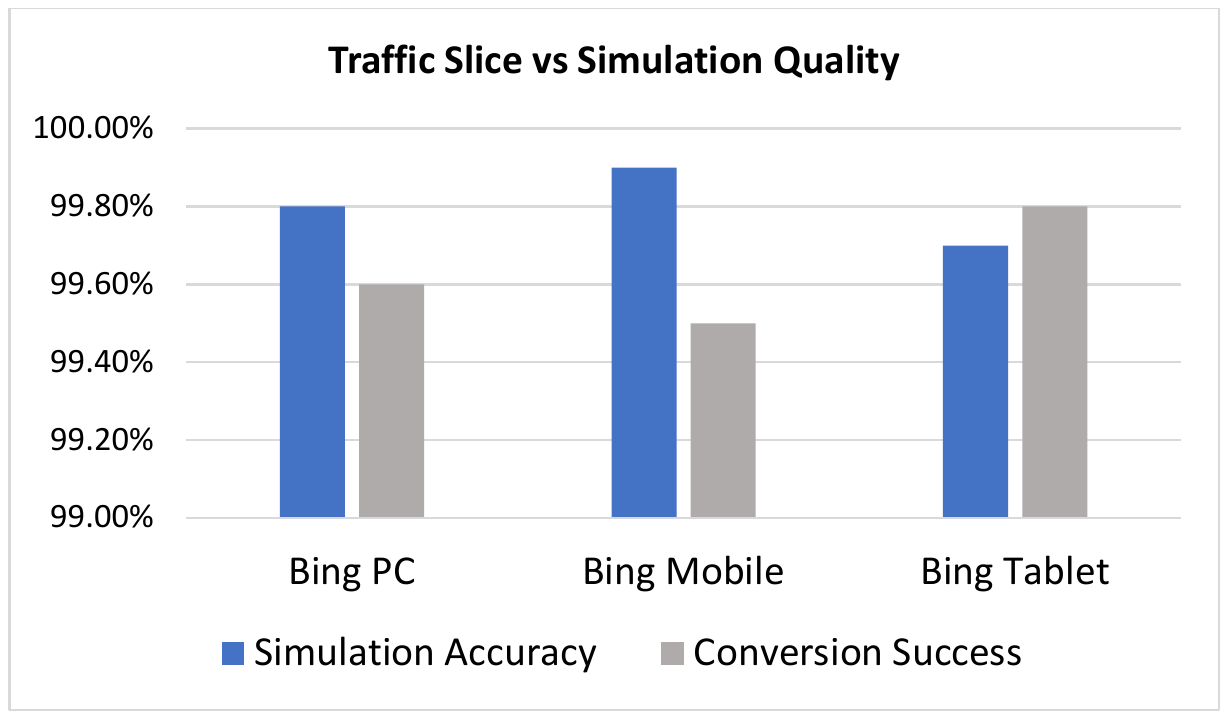}}%
	  \caption{\small Simulation Quality \label{fig:sim-quality}}%
  \end{minipage}%
	\begin{minipage}{0.5\linewidth}%
    \centering
		\subfloat{\includegraphics[width=1.0\textwidth, viewport=130 530 520 705]{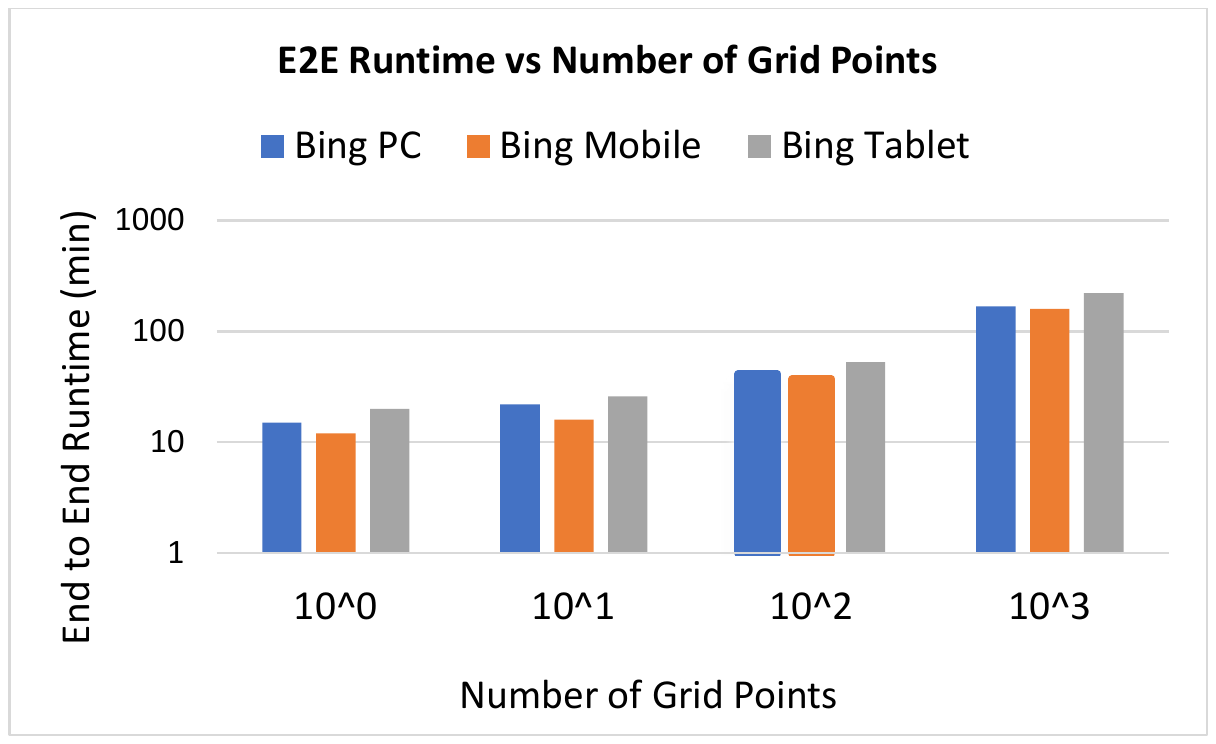}}%
	  \caption{\small Simulation Runtime \label{fig:sim-runtime}}%
  \end{minipage}%
  \end{center}
\end{figure*}

\begin{figure}[h!]
	\centering
		\includegraphics[width=0.5\textwidth, viewport=130 530 520 700]{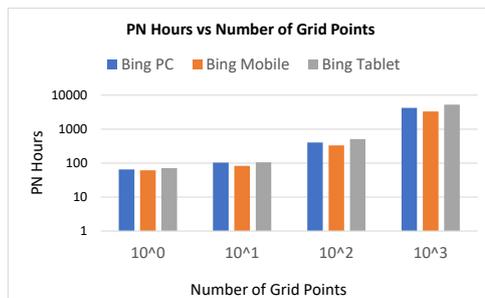}
	\caption{\small Simulation PN Hours \label{fig:sim-pn-hours}}%
\end{figure}

\subsection{Auction Simulation Experiments}
In this part, we analyze the runtime performance and replay quality of the Auction Simulation step. For the experimental purpose, we use the 3 different data samples collected from Bing PC, Bing Mobile and Bing Tablet traffic slices in January 2018. 

In this experiment, we prepare $12$ job configs for $3$ traffic slices with $4$ different grid settings. The performance results only reflects simulation phase of the auction simulation job that is executed on Cosmos platform. The Auction Simulation component runs on $2000$ processing node up to certain number of data samples cross the number of input grid points. Once the number of input grid points and data size increase, we gradually increase the number of processing nodes based on heuristics which is calibrated periodically. In the current experiment, we used 4 different grid size on exponential scale from $10^{0}$ to $10^{3}$. The first 3 data points correspond to simulation run with $2000$ nodes and the grid size of $10^{3}$ corresponds to simulation run with $10000$ processing nodes.

Figure~\ref{fig:sim-quality} presents the simulation quality for this experiment. Simulation accuracy represents the percentage of requests that can be replayed correctly in offline among converted logs. Conversion success represents the percentage of requests which are logged properly and can be converted into the data structure that online library expects. As Figure~\ref{fig:sim-quality} suggests that, both simulation accuracy and conversion success are constantly above $99\%$ which is needed for accurate counterfactual estimation. Figure~\ref{fig:sim-runtime} and Figure~\ref{fig:sim-pn-hours} present runtime performance metrics. As the number of grid points increases, the runtime and PN hours increase too. Another observation is that the tangent of end to end time is not sharp as in total PN hours till some point ($10^{3}$). This implies that as long as there is an available resource for parallelization, the end to end runtime is less impacted compared to the total processing time.

\begin{figure*}[ht!]
  \begin{center}
  \begin{minipage}{0.5\linewidth}%
    \centering
		\subfloat{\includegraphics[width=1.0\textwidth, viewport=130 530 520 705]{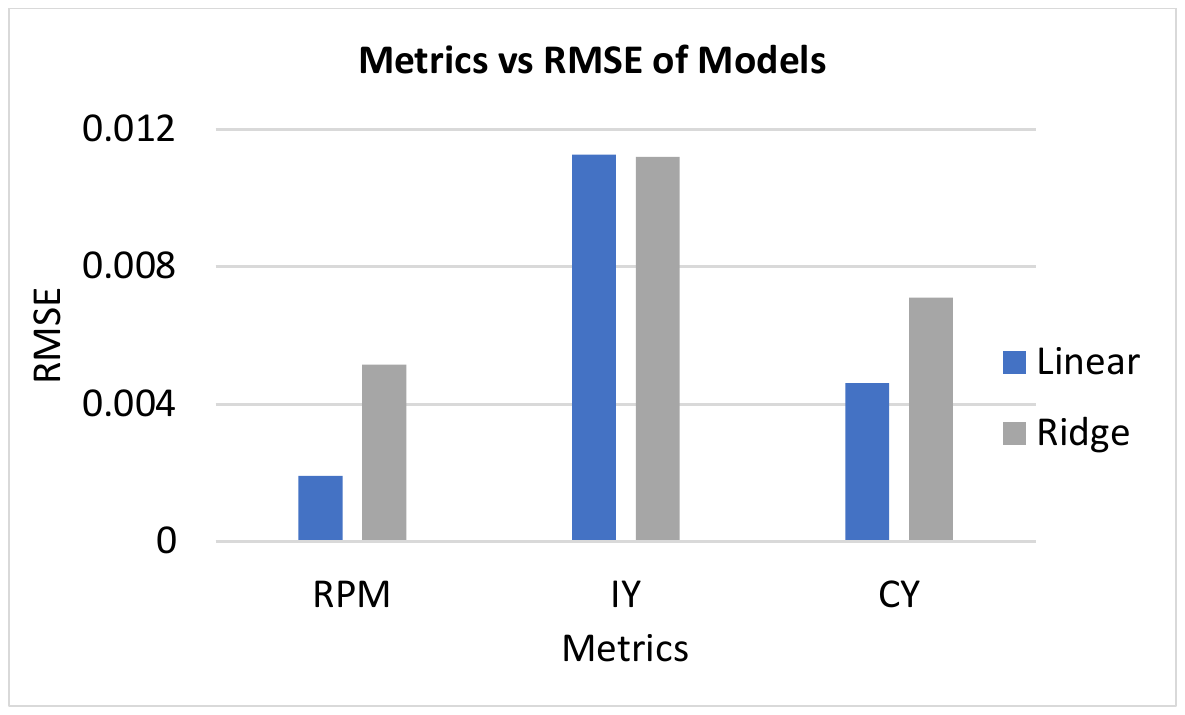}}%
	  \caption{\small Model Accuracy \label{fig:grid-accuracy}}%
  \end{minipage}%
	\begin{minipage}{0.5\linewidth}%
    \centering
		\subfloat{\includegraphics[width=1.0\textwidth, viewport=130 530 520 705]{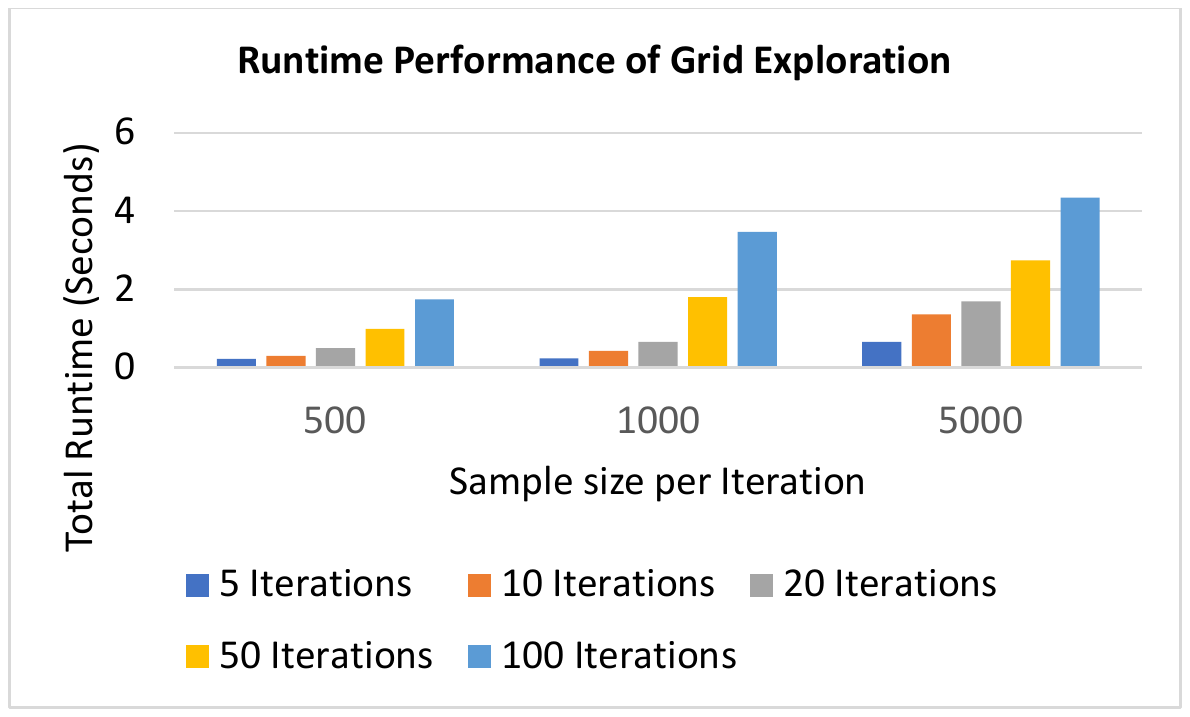}}%
	  \caption{\small Grid Exploration Performance \label{fig:grid-runtime}}%
  \end{minipage}%
  \end{center}
\end{figure*}

\begin{figure}[h!]
	\centering
		\includegraphics[width=0.5\textwidth, viewport=130 530 520 705]{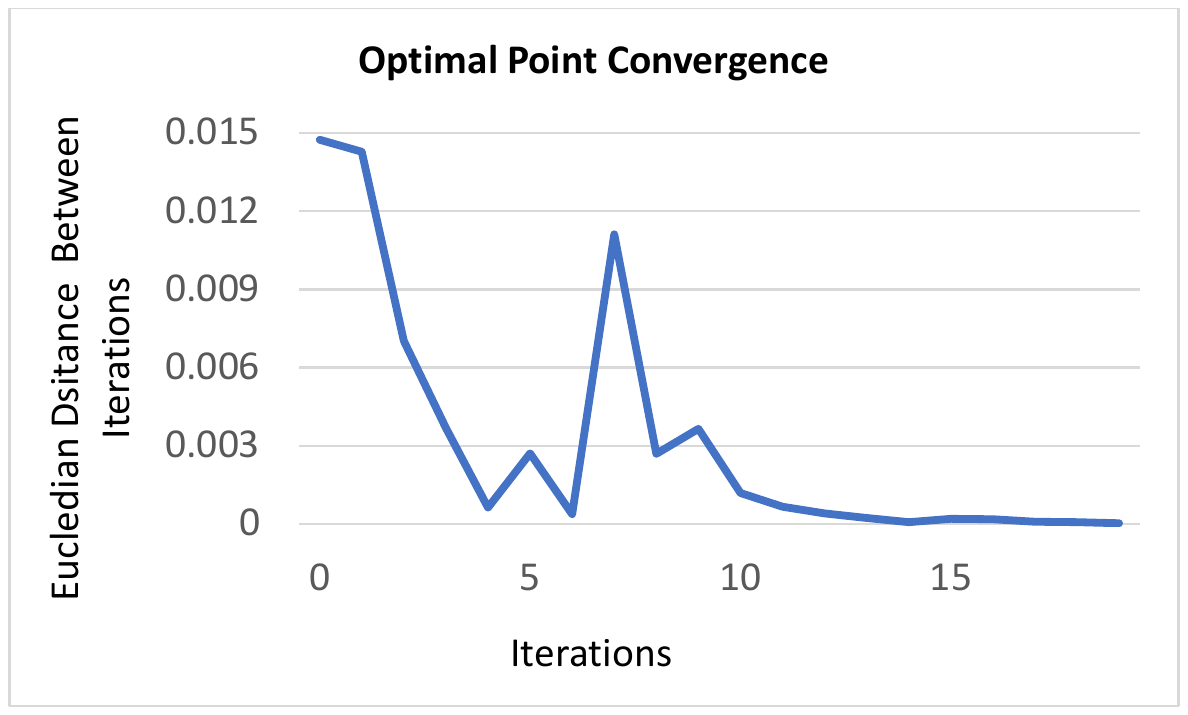}%
	\caption{\small Optimal Point Conversion \label{fig:grid-conversion}}%
\end{figure}

\subsection{Grid Exploration}

For the grid exploration, we use one of the specific tuning scenario for Bing PC slice with $2K$ grid points that spans a week of data in Jan 2018. Through this section, Grid Exploration experiments run in a single machine that has Intel i7 CPU with $4$ cores and $32$ GB RAM. 

In the first experiment, we compared the prediction accuracy of Linear and Ridge regression. For both models, we tried 3 different versions that use linear, quadratic and cubical polynomial feature space. Here, we don't present higher polynomial feature space than cubical since increasing the order of polynomial feature space more causes over fitting and does not contribute prediction accuracy significantly. For reporting purpose, we only compare the most common metrics that are used for tuning decisions: Revenue per mille (RPM), click yield (CY) and impression yield (IY). Figure~\ref{fig:grid-accuracy} presents the Root Mean Square Error (RMSE) comparison of Ridge and Linear regression. In this experiment, we use 3-fold cross validation where we randomly divide data into 3 subset and 2 subset is used for training and rest is used for prediction. Our first finding is that Linear regression model yields better prediction performance compared to Ridge regression. We also discovered that cubical features yields best performance in majority of the tuning scenarios.

In the second and third experiments, we analyze the performance of the optimization algorithm mentioned in Section~\ref{sec:grid}. Figure~\ref{fig:grid-runtime} presents the runtime performance with respect to the number of solutions explored in each step and the number of maximum iterations. As we increase the number of iterations and solution space, the runtime increases linearly. In terms of number of solutions, increasing the solution space in each iteration does not contribute significantly after $5000$ solutions. Another interesting results is that, the grid exploration method with $5000$ solution space converges very quickly after $20$ iterations (Figure~\ref{fig:grid-conversion}). This is indeed a very common behavior that we observed in thousands of tuning jobs where grid size coming from Genie jobs are less than $5K$.

\subsection{Comparison of Policy Estimators}

In this section, we compare counterfactual estimation performance of Genie with Importance Sampling based estimator~\cite{bottou2013} which is one of the most common observational approaches that is utilizing inverse propensity weight estimation. Importance Sampling computes any metric y based on given online distribution of system parameters $P(x)$ over sample of $N$ data points as follows:

\begin{equation}
  Y =  \int_{x}yP(x) \approx \frac{1}{N}{\sum}_{i=1}^{N}y_{i}
\end{equation}

For the counterfactual distribution $P^{*}(x)$ of online parameters, the Counterfactual KPI $Y^{*}$ can be estimated as follows where the $w(x_{i})$ corresponds to ratio of counterfactual distribution over proposal distribution for current sample $x_{i}$:

\begin{equation}
  Y^{*} =  \int_{x}y\frac{P^{*}(x)}{P(x)}P(x) \approx \frac{1}{N}{\sum}_{i=1}^{N}{\frac{P^{*}(x_{i})}{P(x_{i})}}y_{i} = \frac{1}{N}{\sum}_{i=1}^{N}w(x_{i})y_{i}
\end{equation}

Since Importance Sampling needs online randomization of tunable parameter set $x$ under certain distribution of $P(x)$, we choose less risky tuning scenarios that is used for periodic tuning of existing model parameters. For comparison purpose, we choose 5 tuning interval for the same problem between April and May 2018 for Bing PC traffic slice. We also created randomized experiment with Gaussian distribution on certain set of parameters that are used for controlling the same model. After that, we tuned same parameter set periodically 5 times via importance sampling on randomized logs with a certain objective function and a set of k proposal distributions $PD$ = $\{P_{1}^{*}(x),$ $P_{2}^{*}(x)$, \ldots, $P_{k}^{*}(x)\}$. At the end of each tuning period, the best solution from corresponding tuning jobs is applied to treatment traffic under $A/B$ experiment setup for certain amount of testing time. For comparison purpose, we also estimated the KPI impact of selected solutions (selected proposal distribution) for each time interval by using Genie tunings on exactly same data.

\begin{table}[h!]
	\centering
  \caption{Comparison with Importance Sampling}
  \label{tab:comparison}
  \begin{tabular}{ccccc}
    \toprule
    Method&RPM&MLIY&CY&CPC\\\cline{1-5}
		IS (Historical)&1.27\%&0.41\%&0.39\%&1.14\%\\\cline{1-5}
		Genie (Historical)&\textbf{1.16\%}&\textbf{0.32\%}&\textbf{0.37\%}&\textbf{0.93\%}\\\cline{1-5}
		IS (Regression)&0.90\%&0.36\%&\textbf{0.24\%}&0.98\%\\\cline{1-5}
		Genie (Regression)&\textbf{0.88\%}&\textbf{0.25\%}&0.27\%&\textbf{0.66\%}\\
  \bottomrule
\end{tabular}
\end{table}

Table~\ref{tab:comparison} presents comparison of Genie Estimator with Importance Sampling based Estimator. The first and second rows represent the average error from tunings based on data collected from randomized experiment during tuning interval. The error in this context is the distance between predicted KPI delta of $k^{th}$ interval and actual KPI delta of $(k+1)^{th}$ interval that is computed from the $A/B$ experiment. The last two rows represent the average error from regression tunings where we apply same counterfactual on control traffic of $A/B$ tests on each interval (predicted and actual KPIs belong to same interval). Our first observation is that Genie outperforms Importance Sampling estimator on the space (MLIY) and pricing metrics (CPC) on noisy traffic slices since the system layer in causal graph has significant impact on the outcome of these variables. Another interesting observation is that both estimator performs closer for Click Yield and RPM the outcome of which is significantly impacted by user behaviors directly or indirectly. This behavior is indeed pretty similar in many other tunings scenarios where we observe that Genie is more successful on predicting metrics that has more noise due to system related changes like frequent policy updates. Our final observation is that both tunings give better result if they use the same control data for tuning with $A/B$ testing time period which are listed as regression tunings in Table~\ref{tab:comparison}.

\subsection{Tuning New Policies with Genie}

In this section, we present experimental results on tuning new policies with Genie for Bing PC traffic. Tuning new policies is the most risky tuning scenarios since there may not be enough knowledge on hyperparameter space. Thus, creating $A/B$ testing or randomized experiment could be very costly due to cold start problem.

For this particular experiment, we used periodic combo tuning scenario where parameters that control new models from multiple teams are tuned jointly to achieve global optimization goals. To find the best operating point for this tuning task, Genie user submitted more than 6 jobs. After tunings, user published the best operating point and created two online experiment for $A/B$ testing which are randomly sampled versions of the current production Bing PC traffic slice. Then, the first experimental slice is updated to the new operating point by applying optimal parameter settings as a treatment and the second clone is kept as a control slice for $A/B$ testing comparison.

\begin{table}[h!]
  \centering
  \caption{End to End Tuning Experiments}
  \label{tab:end-to-end}
  \begin{tabular}{ccccc}
    \toprule
    Job&Dates&$\Delta$RPM&$\Delta$CY&$\Delta$MLIY\\\cline{1-5}
		Real Delta&01/31-02/02&+1.60\%&+0.14\%&+4.74\%\\\cline{1-5}
		Genie Tuning&01/31-02/02&+1.73\%&+0.07\%&+5.13\%\\\cline{1-5}
	  Genie Tuning&01/24-01/30&+1.81\%&+0.04\%&+5.33\%\\
  \bottomrule
\end{tabular}
\end{table}

For KPI validation purpose, we also submit Genie job on control traffic of $A/B$ test by applying same treatment in simulation for comparison purpose. The results of current experiments are given in Table~\ref{tab:end-to-end} for RPM (Revenue Per Mille), CY (Click Yield), MLIY (Main Line Impression Yield). Our first observation is that Genie results are consistent with real traffic and metrics of interests are directionally correct. Our second observation is that results are getting better when data for real traffic and logs of Genie job are similar and sampled from same dates (row 1 and row 2 in Table~\ref{tab:end-to-end}). When we shift temporal dimension to the past, we started to observe more discrepancy (delta of row 1 and row 3) since user and advertiser behaviors are getting different as training data gets older compared to $A/B$ testing time frame.

\section{Challenges and Lessons Learned}
\label{sec:challenges}

Maintaining large scale counterfactual policy estimator like Genie is a challenging task. In this section, we discuss the major challenges and lessons learned from our previous experiences.

\subsection{Online vs Offline Parity for New Features}

Online and Offline parity is the biggest challenge for maintaining Genie platform. Although online and offline binaries need to include same logic from initial ad selection to the page allocation, there are substantial differences between these two binaries due to the environments that they are living in. In online serving environment, the infrastructure is optimized for reducing end to end latency per user query. In offline environments like Cosmos, the infrastructure is optimized for handling large scale data and reducing the overall time of the costly big data jobs. 

Due to the different optimization goals in the online and offline environments, there are substantial amount of unshared code between Genie binaries and Bing Ads Online Serving binaries. If the unshared code is not managed properly, the offline binaries can have significant amount of regression while replaying the logs to get same result compared to one that was delivered on online serving time. To solve this problem, we isolated all of the algorithmic logic into separate dlls which are exactly same in both online and offline environment. The rest of the environment specific code is living in corresponding wrapper dlls in both environments. After this setup, we marked the methods that are called from wrappers to the shared library. Whenever, there is a code change by any developer for a new feature that touches interface of shared library from online wrapper library, we trigger extra testing and functional gates to make sure that existing functionality is not broken and new functionality is working.

\subsection{Backward Compatibility}
Backward compatibility is another challenge for maintaining the large scale log replay platform. As the codebase evolves, it may not be possible to replay historical logs after some time due to major version changes. On the other hand, majority of the marketplace tunings require most recent historical data and teams need to make sure that online changes do not broke replay functionality in particular time window in the past. In our current design, we do not allow in-place code changes directly to the existing logic in the code. Developers create a new knob (system parameter) and all changes are implemented in the conditional branch that assumes existence of new knob (knob=true). In this setup, the offline log replay simulator can run both old and new logs with their corresponding knobs without any compatibility issue.

\subsection{Increasing Data Size and Search Space}
The Bing Ads Data is growing aggressively in several dimensions including traffic volume, ad data and contextual data like user signals. In addition, the complexity of the algorithmic logic increases significantly which requires tuning more policy parameters. Due to the increasing complexity of the problem space, we continuously monitor the performance of the log replay jobs. The structure of the scope jobs are periodically reviewed and the number of processing nodes for critical steps are calibrated accordingly. For the growing system parameters, tuning users determine the mutually exclusive subsets of tunable parameters to the best of their knowledge and each subset is tuned independently from other subsets.

\subsection{Feature Engineering for Click Prediction}

Last but not least, Feature Engineering also poses challenge for Genie as most of the jobs are still using Bayesian Probit model due to high performance. As mentioned in previous sections, Bayesian Probit model supports only categorical features. Important continuous features like ad pixel height, ad block pixel and position independent click probability score need to be binned properly. Another problem with Probit model is that it does not learn feature importance and feature combination from the data. Although the number of individual features is limited, there are many ways to construct feature combinations. Selecting important individual and combined features costs significant amount of time for us. Switching completely to Tree Boosting can significantly reduce feature engineering efforts. 

\section{Conclusion Remarks}
\label{sec:conclusions}

In this paper, we propose an offline counterfactual policy estimation framework called Genie to optimize the Sponsored Search marketplace. From the experimental results on Bing traffic, we show that Genie can be used to tune completely new policy changes or parameters of existing policies for noisy traffic slices. We also discovered that Genie can yield better results compared to existing observational approaches for metrics the outcome of which depend more on policy changes. For metrics the outcome of which depends on user behavior, Genie can also produce similar KPI results compared to observational approaches like Importance Sampling. Thus, system simulation + user modeling mechanism that Genie is utilizing could be applied to similar problem contexts where training data has bias due to frequent policy updates.

\section{Acknowledgements}
The authors would like thank Emre Kiciman for his feedbacks on improving the content of this paper, Tianji Wang and Qian Yao for their help on running Policy Estimators comparision experiments. Finally, authors would like to thank Rukmini Iyer, Eren Manavoglu, Patrick Jordan, Manish Agrawal, Patryk Zaryjewski, Anton Schwaighofer, Sendill Arunachalam, Pei Jiang, Suyesh Tiwari, Ranjita Naik, Marcin Machura, Urun Dogan, Elon Portugaly, Jitendra Ajmera, Aniruddha Gupta, Dilan Gorur, Debabrata Sengupta, Eugene Larchyk, Tommy Tan, Xiaoliang Ling, Thomas Borchert, many talented scientists and engineers in Microsoft Bing Ads Team for their help on Genie implementation and feedbacks for many features of Genie.

The auction described in the paper is for illustration only and it is not necessarily the one that Bing uses.

\bibliographystyle{plain}
\bibliography{sample-bibliography} 

\end{document}